\documentclass[letterpaper,journal]{IEEEtran}

\usepackage{amsmath,amsfonts,amssymb}
\usepackage{array}
\usepackage[caption=false,font=normalsize,labelfont=sf,textfont=sf]{subfig}

\usepackage{stfloats}
\usepackage{url}
\usepackage{graphicx}
\usepackage{cite}
\usepackage{booktabs}
\usepackage{tabularx}
\usepackage[dvipsnames]{xcolor}
\usepackage{xspace}

\graphicspath{{figures/}}
\definecolor{cuda}{HTML}{4568FF}
\definecolor{acados}{HTML}{F4A261}
\definecolor{casadi}{HTML}{E63946}
\definecolor{sls}{HTML}{4e9a86}
\definecolor{ilqr}{HTML}{264653}
\definecolor{pimpcjulia}{HTML}{ab4e68}
\definecolor{pimpcpytorch}{HTML}{e74a2b}

\newcommand{\cudampc}{\textsc{CudaMPC}\xspace}
\newcommand{\piMPC}{\(\pi\)MPC\xspace}
\newcommand{\swatch}[1]{\raisebox{0.5ex}{\colorbox{#1}{\hspace{0.03em}}}}

\begin{document}

\title{\textsc{CudaMPC}: A GPU-Native Solver for Model Predictive Control}

\author{\IEEEauthorblockN{Babak Akbari and Melissa Greeff}
\thanks{Robora Lab, Department of Electrical and Computer Engineering\\
Queen's University, Kingston, ON, Canada\\
\{babak.akbari, melissa.greeff\}@queensu.ca}}

\maketitle

\begin{abstract}
Model Predictive Control (MPC) delivers constraint-aware control, but its
reliance on online optimization limits its use on systems with fast dynamics,
high-dimensional models, or long horizons. Existing GPU implementations
typically treat the device as a linear-algebra accelerator, leaving the
optimization loop dependent on repeated kernel launches and high-latency memory
transfers. This paper introduces \cudampc, a GPU-native MPC framework that
co-designs the optimization algorithm, execution model, and memory architecture
for CUDA hardware. \cudampc pairs a parallel-in-horizon alternating direction
method of multipliers (ADMM) splitting with a fused CUDA kernel that runs the
entire iterative solve on the device. Intermediate optimization variables stay
in low-latency, on-chip shared memory, and a localized atomic-flag protocol
synchronizes only adjacent horizon blocks, minimizing host intervention,
kernel-dispatch overhead, and global-memory traffic. Across six nonlinear
robotics benchmarks spanning increasing state dimension and constraint density,
\cudampc sustains real-time rates at horizons one to two orders of magnitude
longer than CPU solvers: it solves an optimization-based collision-avoidance
parking problem with \(100\,\mathrm{s}\) of lookahead within a
\(0.1\,\mathrm{s}\) sampling interval, and is the only solver evaluated that
achieves both real-time execution and collision-free coordination for a
centralized 10-agent swarm, where acados and CasADi return no feasible solution
and require \(3.5\)~s and \(4.5\)~s per solve. Against tensor-framework
implementations of the same ADMM splitting, the fused kernel is up to
\(965\times\) faster.
\end{abstract}

\section{Introduction}

Model Predictive Control (MPC) is among the most established advanced control
methodologies, owing to its ability to explicitly enforce state and input
constraints while optimizing multivariable performance over a receding
horizon~\cite{mayne2000constrained,qin2003survey}. Advances in embedded
optimization have extended it from process control to fast, nonlinear systems:
autonomous vehicles~\cite{falcone2007predictive,liniger2015optimization}, mobile
and aerial robots~\cite{ostafew2016learning,greeff2018flatness}, agile quadrotor
flight~\cite{romero2022model}, and aerospace flight
control~\cite{keviczky2006receding}.

Despite strong theoretical foundations~\cite{zeilinger2014soft}, the online
optimization required by MPC remains a significant computational burden, and one
that binds hardest on exactly these fast robotic platforms. At each sampling
instant MPC solves a constrained optimization problem---a quadratic program (QP)
for linear systems, or a nonlinear program (NLP) otherwise---whose cost grows
rapidly with the state and input dimensions, the number of constraints, and the
horizon length. For fast-sampled systems, CPU-based implementations may therefore
be unable to compute a sufficiently accurate control action within the sampling
interval, forcing compromises in horizon length, model fidelity, or optimization
accuracy~\cite{zeilinger2014realtime}. CPU-oriented frameworks such as
acados~\cite{Verschueren2021} and CasADi~\cite{Andersson2019} combine
structure-exploiting formulations with highly optimized QP and NLP solvers, but
their cost still grows with the horizon and model dimensions, motivating
approaches that expose greater parallelism.

\begin{figure}[t]
\centering
\includegraphics[width=\linewidth]{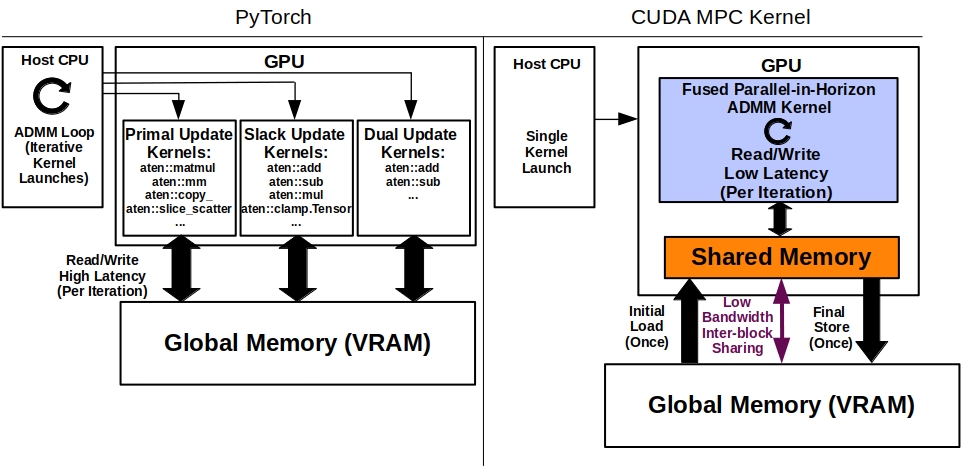}
\caption{\cudampc is designed \emph{for} the GPU, not merely executed on it.
Rather than using the GPU only as a linear-algebra accelerator through a tensor
framework (e.g., PyTorch), \cudampc tailors the complete MPC solution process to
the CUDA architecture, combining parallel-in-horizon ADMM, fully GPU-native
execution, and block-resident computation in low-latency shared memory.}
\label{fig:design_overview}
\end{figure}

With thousands of parallel cores, NVIDIA's Compute Unified Device Architecture
(CUDA) is a natural platform for such parallelism, but adoption in real-time
control has been hindered by the sequential structure of traditional
optimal-control algorithms and by CPU--GPU communication latency. Tensor
frameworks such as PyTorch and JAX~\cite{Paszke2019,Frostig2018} provide
convenient GPU-accelerated linear algebra and automatic differentiation
(Fig.~\ref{fig:design_overview}). However, when iterative optimization is
expressed as a sequence of framework-level operations rather than a single fused
device program, it incurs repeated kernel launches and global-memory traffic,
which at high-frequency MPC rates can dominate the solve time and erode the
parallel advantage of the GPU.

This paper introduces \cudampc, a GPU-native MPC framework co-designed at the
algorithmic, execution, and memory levels (Fig.~\ref{fig:design_overview}).
First, it maps the parallel-in-horizon alternating direction method of
multipliers (ADMM) formulation of~\cite{wu2026pi} onto the CUDA thread-block
hierarchy, distributing the horizon across streaming multiprocessors (SMs) and coupling
them through a localized atomic-flag protocol that synchronizes only adjacent
blocks rather than the whole grid. Second, the complete optimization loop
executes natively on the GPU, without returning control to the host between ADMM
iterations. Third, optimization variables and intermediate computations remain in
low-latency, on-chip shared memory throughout the solve. This input/output
(I/O)-aware design follows the principle underlying
FlashAttention~\cite{dao2022flashattention}: performance improves not only by
increasing arithmetic parallelism, but also by minimizing transfers between
on-chip and higher-latency global memory.

The contributions of this paper are:\footnote{Source code will be released upon
acceptance.}
\begin{itemize}
\item \textit{The \cudampc framework:} a GPU-native MPC architecture executing
parallel-in-horizon ADMM within a single, shared-memory-resident CUDA kernel.
\item \textit{A neighbor-local synchronization protocol} replacing grid-wide
barriers with pairwise atomic flags, so coordination cost is independent of the
number of horizon blocks.
\item \textit{A benchmark study} against state-of-the-art CPU and GPU solvers on
six nonlinear robotics problems, showing that the resulting increase in
real-time-feasible horizon yields closed-loop capability the CPU baselines cannot
achieve at any horizon they can run.
\end{itemize}

\section{Related Work}

Work related to \cudampc falls into two strands. In the first, the
methodological advance lies in the control formulation---improved models,
higher-fidelity actuator dynamics, task-specific objectives, or more expressive
constraints---while the optimization is delegated to established libraries such
as acados or CasADi. Representative examples include aggressive quadrotor control
using DQ-NMPC~\cite{Recalde2025DQNMPC}, low-level-dynamics-aware
NMPC~\cite{Gupta2025LoLNMPC}, UAV landing on a moving surface
vessel~\cite{Prochazka2024BoatMPC}, and predictive control of surface vessels
under wave disturbances~\cite{Jiang2025USV,Jenkins2024USV}. These advance the
modeling and decision-making layers of MPC rather than redesigning the solver for
the target hardware.

The second strand exposes greater parallelism by redesigning parts of the MPC
optimization for GPUs, through custom CUDA implementations, QP solvers expressed
in machine-learning frameworks, and parallel-in-time methods based on associative
scans~\cite{adabag2024mpcgpu,bishop2024relu,iacob2025parallel,wu2026pi,fang2026safe}.
Unlike prior approaches that accelerate individual numerical operations or map an
existing solver onto a GPU framework, \cudampc restructures the MPC algorithm
around the CUDA execution and memory hierarchy.

\subsection{CPU-Based MPC Solvers}

Real-time nonlinear MPC has traditionally relied on highly optimized numerical
methods executed on CPUs. CasADi~\cite{Andersson2019} provides symbolic modeling,
automatic differentiation, and interfaces to nonlinear programming solvers such
as Ipopt~\cite{wachter2006implementation}. acados~\cite{Verschueren2021} instead
exploits the multistage structure of optimal control problems through tailored
Sequential Quadratic Programming (SQP) and Real-Time Iteration (RTI) schemes; its
structured QP solvers, including HPIPM~\cite{frison2020hpipm}, use block-sparse
factorizations and Riccati-type recursions to achieve real-time performance.
These frameworks have enabled increasingly sophisticated onboard controllers: for
example, DQ-NMPC employs acados, SQP-RTI, and HPIPM for aggressive quadrotor
flight at rates up to \(190\,\mathrm{Hz}\) on an NVIDIA Jetson Orin
NX~\cite{Recalde2025DQNMPC}.

The efficiency of these methods follows from exploiting the structured sparsity
of the optimal control problem. Their factorization stages, forward--backward
sweeps, and Riccati-style recursions nevertheless retain stagewise dependencies
across the horizon, so cost grows with horizon length and problem dimension.
First-order methods offer an alternative with simpler iteration steps.
General-purpose ADMM solvers such as OSQP~\cite{stellato2018osqp} are effective
for structured QPs, and TinyMPC~\cite{nguyen2024tinympc} caches problem
factorizations to run linear MPC on resource-constrained microcontrollers. These
methods remain principally CPU-oriented and do not target horizon-parallel
execution of large nonlinear MPC problems.

\subsection{GPU-Accelerated MPC}

The most widely deployed GPU-based predictive controller in robotics is
sampling-based: model predictive path integral (MPPI)
control~\cite{williams2017mppi,williams2017information} rolls out thousands of
perturbed input sequences in parallel, which maps naturally onto GPU hardware and
requires no derivatives. MPPI is complementary rather than a competing baseline:
it enforces constraints only through cost penalties, provides no feasibility
certificate, and its sample complexity grows with input dimension and
horizon---precisely the regime targeted here. \cudampc instead pursues
gradient-based, constraint-enforcing MPC.

Among gradient-based methods, recent GPU-accelerated MPC differs in where
parallelism enters the optimization pipeline. One class accelerates selected
numerical kernels within otherwise conventional trajectory-optimization methods.
MPCGPU~\cite{adabag2024mpcgpu}, for instance, uses custom GPU kernels for
Schur-complement construction, linear-system solution, and line search. Although
hardware-aware, its parallelism is concentrated in the Newton-system solve,
whereas \cudampc reformulates the MPC subproblem itself to expose parallelism
across the horizon and handles state and input constraints through ADMM
projections.

A second class maps optimization iterations onto GPU-enabled tensor frameworks
such as PyTorch or JAX. ReLU-QP~\cite{bishop2024relu} expresses ADMM iterations
as a weight-tied ReLU network, and implementations of \piMPC~\cite{wu2026pi} use
a parallel-in-horizon ADMM formulation admitting independent stagewise updates.
Such frameworks provide convenient GPU execution, but implementations written as
sequences of framework-level operations incur repeated kernel launches and
global-memory accesses. \cudampc instead executes the complete ADMM loop within a
fused CUDA kernel and retains intermediate primal and dual variables in
low-latency, on-chip shared memory.

A third class parallelizes the recursive structure of optimal control using
associative scans. Temporal dynamic programming~\cite{sarkka2022temporal},
Parallel-in-Time Newton~\cite{iacob2025parallel}, Primal-Dual
iLQR~\cite{amatucci2025primal}, and GPU-SLS~\cite{fang2026safe} recast Riccati,
dynamic-programming, or KKT recursions as tree-structured prefix operations,
reducing parallel depth from \(\mathcal{O}(N)\) to \(\mathcal{O}(\log N)\). These
methods still require horizon-wide upsweep and downsweep passes, with the
associated storage, synchronization, and communication of intermediate scan
variables.

\cudampc exploits temporal parallelism differently: rather than parallelizing a
global forward--backward recursion, it uses ADMM splitting to produce local
stagewise updates coupled through neighboring-block consensus. Each CUDA block
processes a contiguous sub-horizon in on-chip shared memory, exchanging only
boundary variables with its neighbors. This avoids repeated horizon-wide scans
and keeps the complete ADMM iteration inside a single fused kernel.

\section{Problem Statement}

We consider the constrained finite-horizon optimal control problem underlying
nonlinear MPC:
\begin{equation}
\label{eq:nmpc_ocp}
\begin{aligned}
    \min_{\mathbf{x}_{0:N},\,\mathbf{u}_{0:N-1}}
    \quad &
    \sum_{k=0}^{N-1}
    \ell(\mathbf{x}_k,\mathbf{u}_k)
    + \ell_f(\mathbf{x}_N)
    \\
    \text{subject to}
    \quad &
    \mathbf{x}_0 = \mathbf{x}(t),
    \\
    &
    \mathbf{x}_{k+1}
    =
    \mathbf{f}(\mathbf{x}_k,\mathbf{u}_k),
    \qquad k=0,\ldots,N-1,
    \\
    &
    \mathbf{x}_k \in \mathcal{X},
    \qquad k=0,\ldots,N,
    \\
    &
    \mathbf{u}_k \in \mathcal{U},
    \qquad k=0,\ldots,N-1 .
\end{aligned}
\end{equation}
Here \(\mathbf{x}_k\in\mathbb{R}^{n_x}\) and \(\mathbf{u}_k\in\mathbb{R}^{n_u}\)
are the predicted state and input at stage \(k\), \(\mathbf{x}(t)\) is the
measured state,
\(\mathbf{f}:\mathbb{R}^{n_x}\times\mathbb{R}^{n_u}\rightarrow\mathbb{R}^{n_x}\)
are the nonlinear discrete-time dynamics, \(N\) is the horizon length, and
\(\ell,\ell_f\) are the stage and terminal costs. At each sampling instant the
first element of the resulting control sequence is applied to the system.

\section{Background}
\label{sec:background}

\subsection{Sequential Convexification}

We solve the nonlinear MPC problem using Sequential Convex Programming
(SCP)~\cite{Mao2016SCvx,Mao2017SCvx}, which at each MPC update constructs convex
subproblems by linearizing the nonlinear dynamics---and, when necessary,
nonconvex constraints---about the trajectory
\(\{\mathbf{x}_k^{j},\mathbf{u}_k^{j}\}_{k=0}^{N-1}\) obtained at the preceding
outer iteration \(j\), \(j=0,\ldots,j_{\max}-1\). A first-order approximation of
the dynamics gives
\begin{equation}
\label{eq:scp_dynamics}
    \mathbf{x}_{k+1}^{j+1}
    =
    \mathbf{A}_k^j\mathbf{x}_k^{j+1}
    +
    \mathbf{B}_k^j\mathbf{u}_k^{j+1}
    +
    \mathbf{e}_k^j,
\end{equation}
where
\begin{align}
\mathbf{A}_k^j
&=
\left.
\frac{\partial \mathbf{f}}{\partial \mathbf{x}}
\right|_{(\mathbf{x}_k^j,\mathbf{u}_k^j)},
&
\mathbf{B}_k^j
&=
\left.
\frac{\partial \mathbf{f}}{\partial \mathbf{u}}
\right|_{(\mathbf{x}_k^j,\mathbf{u}_k^j)},
\nonumber\\
\mathbf{e}_k^j
&=
\mathbf{f}(\mathbf{x}_k^j,\mathbf{u}_k^j)
-
\mathbf{A}_k^j\mathbf{x}_k^j
-
\mathbf{B}_k^j\mathbf{u}_k^j.
\nonumber
\end{align}
Nonlinear state and input constraints, when present, are likewise replaced by
stagewise polyhedral approximations
\begin{align}
    \mathcal{X}_k^j
    &=
    \left\{
    \mathbf{x}\mid
    \mathbf{C}_k^j\mathbf{x}+\mathbf{c}_k^j\leq\mathbf{0}
    \right\},
    \label{eq:poly_x}                                                \\
    \mathcal{U}_k^j
    &=
    \left\{
    \mathbf{u}\mid
    \mathbf{D}_k^j\mathbf{u}+\mathbf{d}_k^j\leq\mathbf{0}
    \right\}.
    \label{eq:poly_u}
\end{align}
SCP iterations continue until a prescribed convergence tolerance is satisfied or
the maximum number of iterations \(j_{\max}\) is reached.

\subsection{Incremental-Input Formulation}

Following velocity-form MPC formulations~\cite{Sokoler2014InputMPC,wu2026pi},
define \(\Delta\mathbf{u}_k=\mathbf{u}_k-\mathbf{u}_{k-1}\) and the augmented
state
\begin{equation}
\label{eq:augmented_dynamics}
\bar{\mathbf{x}}_k=
\begin{bmatrix}
\mathbf{x}_k\\
\mathbf{u}_{k-1}
\end{bmatrix},
\qquad
\bar{\mathbf{x}}_{k+1}
=
\bar{\mathbf{A}}_k^j\bar{\mathbf{x}}_k
+
\bar{\mathbf{B}}_k^j\Delta\mathbf{u}_k
+
\bar{\mathbf{e}}_k^j,
\end{equation}
where
\[
\bar{\mathbf{A}}_k^j=
\begin{bmatrix}
\mathbf{A}_k^j & \mathbf{B}_k^j\\
\mathbf{0} & \mathbf{I}
\end{bmatrix},
\quad
\bar{\mathbf{B}}_k^j=
\begin{bmatrix}
\mathbf{B}_k^j\\
\mathbf{I}
\end{bmatrix},
\quad
\bar{\mathbf{e}}_k^j=
\begin{bmatrix}
\mathbf{e}_k^j\\
\mathbf{0}
\end{bmatrix},
\]
and \(\bar{n}_x=n_x+n_u\) denotes the augmented state dimension. This form
directly accommodates input-rate penalties and constraints and supports the
parallel-in-horizon splitting of~\cite{wu2026pi}. The corresponding augmented
stagewise feasible set is
\(\bar{\mathcal{X}}_k^j=\mathcal{X}_k^j\times\mathcal{U}_{k-1}^j\), onto which
\cudampc projects during the consensus update.

\subsection{Consensus-Split Convex Subproblem}

To expose parallelism across the horizon, we introduce local copies
\(\mathbf{z}_{k+1}\) and \(\mathbf{v}_k\) and apply an ADMM consensus
splitting~\cite{Boyd2011ADMM,Rey2021ADMM}. At SCP iteration \(j\), the convex
subproblem is
\begin{equation}
\label{eq:consensus_qp}
\begin{aligned}
\min_{\bar{\mathbf{x}},\Delta\mathbf{u},\mathbf{z},\mathbf{v}}
\quad &
\sum_{k=0}^{N-1}
\left[
\ell_k^{\mathrm{QP}}
\left(\bar{\mathbf{x}}_{k+1}^{j+1},
\Delta\mathbf{u}_k^{j+1}\right)
+
\iota_{\bar{\mathcal{X}}_{k+1}^{j}}
\left(\mathbf{z}_{k+1}^{j+1}\right)
\right]
\\
\mathrm{s.t.}\quad
&
\bar{\mathbf{x}}_{k+1}^{j+1}
=\mathbf{z}_{k+1}^{j+1},
\qquad
\bar{\mathbf{B}}_k^j\Delta\mathbf{u}_k^{j+1}
=\mathbf{v}_k^{j+1},
\\
&
\mathbf{z}_{k+1}^{j+1}
=
\bar{\mathbf{A}}_k^j\bar{\mathbf{x}}_k^{j+1}
+\mathbf{v}_k^{j+1}
+\bar{\mathbf{e}}_k^j,
\qquad k=0,\ldots,N-1,
\\
&
\bar{\mathbf{x}}_0=
\begin{bmatrix}
\mathbf{x}(t)^\top &
\mathbf{u}(t-1)^\top
\end{bmatrix}^{\top},
\end{aligned}
\end{equation}
with stage cost
\begin{equation}
\label{eq:convex_stage_cost}
\ell_k^{\mathrm{QP}}
=
\frac{1}{2}\bar{\mathbf{x}}_{k+1}^{\top}
\bar{\mathbf{Q}}_k\bar{\mathbf{x}}_{k+1}
-\bar{\mathbf{q}}_k^{\top}\bar{\mathbf{x}}_{k+1}
+\frac{1}{2}\Delta\mathbf{u}_k^{\top}
\mathbf{R}_k\Delta\mathbf{u}_k,
\end{equation}
where \(\bar{\mathbf{Q}}_k,\mathbf{R}_k\succeq0\) and \(\iota_{\mathcal C}\) is
the indicator function of the convex set \(\mathcal{C}\), whose proximal operator
is the projection \(\Pi_{\mathcal C}\). This splitting separates quadratic
updates, projections, and dynamic consensus across stages, and follows the \piMPC
formulation~\cite{wu2026pi}. Here \(\Pi_{\bar{\mathcal{X}}}\) projects onto the
polyhedral convex sets~\eqref{eq:poly_x}--\eqref{eq:poly_u} obtained by SCP
linearization, rather than the box constraints of a linear-MPC formulation, so
nonlinear dynamics and nonconvex constraints such as OBCA and pairwise collision
avoidance are admitted through the outer SCP loop.

\begin{figure}[t]
    \centering
    \includegraphics[width=\linewidth]{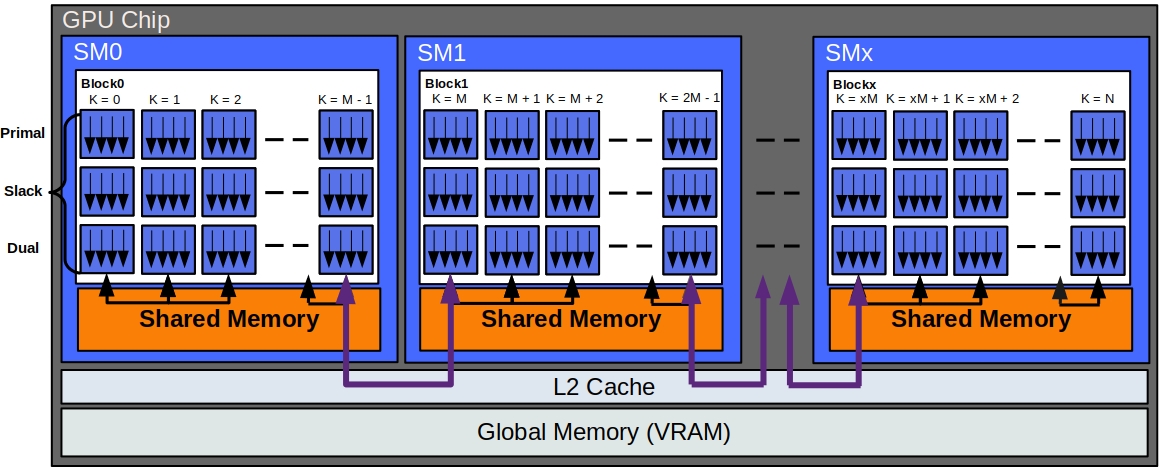}
    \caption{The prediction horizon \(N\) is partitioned into sub-horizons of
length \(M\), distributed across CUDA blocks and GPU streaming multiprocessors
(SMs). Within each block, threads are arranged in an \(M\times\bar{n}_x\) layout,
with one row per prediction stage, and update the primal, consensus, and dual
variables in low-latency shared memory. Only boundary variables are exchanged
between adjacent blocks through global memory, with atomic synchronization
enforcing dynamic consistency across the horizon.}
    \label{fig:gpu_memory_distribution}
\end{figure}

\section{Methodology}

\cudampc is organized around three tightly coupled design layers, described in
turn below. At the \emph{algorithmic} level, a parallel-in-horizon ADMM splitting
of~\eqref{eq:consensus_qp} exposes independent computations across the horizon.
At the \emph{execution} level, these map onto the CUDA thread-block hierarchy and
the complete ADMM iteration is fused into a GPU-native kernel. At the
\emph{memory} level, each block keeps its local optimization variables in
low-latency, on-chip shared memory.

\subsection{Parallel-in-Horizon ADMM}
\label{sec:pih_admm}

For each fixed SCP iteration, the primal updates,
constraint projections, and dual updates of~\eqref{eq:consensus_qp} are evaluated
concurrently over
\(k=0,\ldots,N-1\). Suppressing the SCP index \(j\), we associate the scaled dual
variables \(\boldsymbol{\theta}_k\), \(\boldsymbol{\beta}_k\), and
\(\boldsymbol{\lambda}_k\) with the constraints
\[
\bar{\mathbf{x}}_{k+1}=\mathbf{z}_{k+1},\qquad
\bar{\mathbf{B}}_k\Delta\mathbf{u}_k=\mathbf{v}_k,\qquad
\mathbf{z}_{k+1}
=\bar{\mathbf{A}}_k\bar{\mathbf{x}}_k
+\mathbf{v}_k+\bar{\mathbf{e}}_k,
\]
respectively. For a fixed linearization, define
\begin{subequations}
\begin{align}
\mathbf{J}_k
&:=
\left(\bar{\mathbf{B}}_k^\top\bar{\mathbf{B}}_k\right)^{-1}
\bar{\mathbf{B}}_k^\top,
\label{eq:admm_J}\\
\mathbf{H}_k
&:=
\begin{cases}
\left(\bar{\mathbf{Q}}_k+\rho\mathbf{I}\right)^{-1},
& k=N-1,\\
\left(
\bar{\mathbf{Q}}_k+\rho\mathbf{I}
+\rho\bar{\mathbf{A}}_{k+1}^{\top}\bar{\mathbf{A}}_{k+1}
\right)^{-1},
& k<N-1,
\end{cases}
\label{eq:admm_H}
\end{align}
\end{subequations}
with penalty parameter \(\rho>0\). These matrices are computed once per SCP
linearization and reused across the ADMM iterations.

At iteration \(i+1\), the primal updates are
\begin{subequations}
\label{eq:admm_primal}
\begin{align}
\Delta\mathbf{u}_k^{i+1}
&=
\mathbf{J}_k
\left(\mathbf{v}_k^i-\boldsymbol{\beta}_k^i\right),
\label{eq:admm_du}\\
\bar{\mathbf{x}}_{k+1}^{i+1}
&=
\mathbf{H}_k\mathbf{h}_k^i,
\label{eq:admm_x}
\end{align}
\end{subequations}
with
\begin{subequations}
\label{eq:admm_h}
\begin{align}
\mathbf{r}_{k+1}^i
&:=
\mathbf{z}_{k+2}^i-\mathbf{v}_{k+1}^i
-\bar{\mathbf{e}}_{k+1}
+\boldsymbol{\lambda}_{k+1}^i,
\label{eq:admm_neighbor_residual}\\
\mathbf{h}_k^i
&:=
\begin{cases}
\bar{\mathbf{q}}_k
+\rho(\mathbf{z}_{k+1}^i-\boldsymbol{\theta}_k^i),
& k=N-1,\\
\bar{\mathbf{q}}_k
+\rho\!\left(
\mathbf{z}_{k+1}^i-\boldsymbol{\theta}_k^i
+\bar{\mathbf{A}}_{k+1}^{\top}\mathbf{r}_{k+1}^i
\right),
& k<N-1.
\end{cases}
\label{eq:admm_h_update}
\end{align}
\end{subequations}
These updates are stagewise independent because neighboring quantities are taken
from iteration \(i\).

The consensus quantities are
\begin{subequations}
\begin{align}
\boldsymbol{\gamma}_k^{i+1}
&=
\bar{\mathbf{B}}_k\Delta\mathbf{u}_k^{i+1}
+\boldsymbol{\beta}_k^i
-\bar{\mathbf{A}}_k\bar{\mathbf{x}}_k^{i+1}
-\bar{\mathbf{e}}_k
+\boldsymbol{\lambda}_k^i,
\label{eq:admm_gamma}\\
\boldsymbol{\eta}_k^{i+1}
&=
\bar{\mathbf{x}}_{k+1}^{i+1}
+\boldsymbol{\theta}_k^i
+\bar{\mathbf{A}}_k\bar{\mathbf{x}}_k^{i+1}
+\bar{\mathbf{e}}_k
-\boldsymbol{\lambda}_k^i,
\label{eq:admm_eta}
\end{align}
\end{subequations}
where \(\bar{\mathbf{x}}_0\) is fixed by the current state and previous input.
The projection and consensus updates are
\begin{subequations}
\label{eq:admm_consensus}
\begin{align}
\mathbf{z}_{k+1}^{i+1}
&=
\Pi_{\bar{\mathcal{X}}_{k+1}}
\left(
\frac{2\boldsymbol{\eta}_k^{i+1}
+\boldsymbol{\gamma}_k^{i+1}}{3}
\right),
\label{eq:admm_z}\\
\mathbf{v}_k^{i+1}
&=
\frac{1}{2}
\left(
\mathbf{z}_{k+1}^{i+1}
+\boldsymbol{\gamma}_k^{i+1}
\right),
\label{eq:admm_v}
\end{align}
\end{subequations}
and the scaled dual variables satisfy
\begin{subequations}
\label{eq:admm_dual}
\begin{align}
\boldsymbol{\theta}_k^{i+1}
&=
\boldsymbol{\theta}_k^i
+\bar{\mathbf{x}}_{k+1}^{i+1}
-\mathbf{z}_{k+1}^{i+1},\\
\boldsymbol{\beta}_k^{i+1}
&=
\boldsymbol{\beta}_k^i
+\bar{\mathbf{B}}_k\Delta\mathbf{u}_k^{i+1}
-\mathbf{v}_k^{i+1},\\
\boldsymbol{\lambda}_k^{i+1}
&=
\boldsymbol{\lambda}_k^i
+\mathbf{z}_{k+1}^{i+1}
-\bar{\mathbf{A}}_k\bar{\mathbf{x}}_k^{i+1}
-\mathbf{v}_k^{i+1}
-\bar{\mathbf{e}}_k.
\end{align}
\end{subequations}

Equations~\eqref{eq:admm_primal}--\eqref{eq:admm_dual} are evaluated in parallel
over \(k\). Their local and nearest-neighbor dependencies are what enable the
CUDA block mapping of Sec.~\ref{sec:cuda_execution}.

\subsection{GPU-Native CUDA Execution}
\label{sec:cuda_execution}

\cudampc maps the prediction horizon onto CUDA thread blocks, each assigned a
contiguous sub-horizon of \(M\) stages (Fig.~\ref{fig:gpu_memory_distribution}).
Threads within a block cooperatively evaluate the primal, consensus, and dual
updates. Rather than launching a separate kernel per ADMM operation, the complete
inner loop is fused into a single custom kernel that runs all prescribed
iterations without returning control to the host, eliminating repeated dispatch
and keeping intermediate variables resident on the device.

At the outer MPC level, CUDA Graphs~\cite{nvidia2026cuda} capture the recurring
sequence of trajectory shifting, model linearization, matrix preparation, ADMM
solution, and control extraction; subsequent MPC updates replay this graph with
minimal host-side overhead. CUDA Graphs thus reduce dispatch \emph{between} outer
solver stages, while kernel fusion eliminates dispatch \emph{within} the ADMM
solve.

The grid size is chosen within the cooperative-launch residency limit and
launched with \verb|cudaLaunchCooperativeKernel|, so that all solver blocks are
concurrently resident. Within a block, stages exchange data through shared memory
and synchronize using \verb|__syncthreads()|. Because shared memory is
block-private, only boundary variables are exchanged between adjacent
sub-horizons through global memory.

\paragraph*{Neighbor-local synchronization} A grid-wide \verb|grid.sync()|
barrier after every ADMM sub-step would force all \(B\) blocks to the slowest
block twice per iteration, so its cost grows with the horizon even though the
dependencies in \eqref{eq:admm_primal}--\eqref{eq:admm_dual} are only
nearest-neighbor. \cudampc therefore replaces the global barrier with pairwise
atomic flags in the style of~\cite{xiao2010inter}: 
each block increments a shared stage counter twice per iteration, once after the primal update and once after the consensus update, and spins only on the counters of its immediate spatial neighbors. This makes coordination cost independent of \(B\). Because boundary states are exchanged just prior to their respective synchronization barriers (primal variables propagating forward, and consensus variables propagating backward), the scheme admits a bounded drift of one stage: a block may consume a neighbor's boundary value from iteration \(i-1\) rather than \(i\). The scheme is thus a
\emph{partially asynchronous} ADMM with delay bound one, for which convergence
under bounded delay is established in~\cite{chang2016async}, rather than the
synchronous iteration of~\cite{wu2026pi}.

\subsection{Memory-I/O-Aware Block Residency and Resource Allocation}
\label{sec:memory_design}

Following the I/O-aware principle of FlashAttention~\cite{dao2022flashattention},
\cudampc reduces repeated movement of intermediate iterates between global and
on-chip memory. At kernel initialization, each block cooperatively loads the
matrices and trajectory variables for its sub-horizon into a shared-memory
workspace. The primal variables \(\bar{\mathbf{x}},\mathbf{z},\mathbf{v}\), dual
variables \(\boldsymbol{\theta},\boldsymbol{\beta},\boldsymbol{\lambda}\), and
intermediate ADMM quantities remain resident throughout the solve; global memory
is used only for the initial load, the final output, and the boundary exchange.
Each block deliberately consumes up to the available shared-memory budget to keep
the largest possible sub-horizon on-chip, which on the evaluated GPU limits
residency to one block per SM---trading occupancy for persistent on-chip storage.

The local horizon length follows from the thread and shared-memory limits of the
target device. Let \(T_{\max}\) be the maximum threads per block, \(S_{\max}\)
the available shared memory per block, \(\bar{n}_x\) the augmented state
dimension, and \(S_{\mathrm{step}}\) the shared-memory footprint per stage (which comprises the local state trajectories, consensus variables, and cached system matrices). Then
\begin{align}
M_{\mathrm{threads}}
&=
\left\lfloor
\frac{T_{\max}}{\bar{n}_x}
\right\rfloor,
\label{eq:m_threads}
\\
M_{\mathrm{memory}}
&=
\left\lfloor
\frac{S_{\max}}{S_{\mathrm{step}}}
\right\rfloor,
\label{eq:m_memory}
\end{align}
and the stages per block and total number of blocks are
\begin{align}
M
&=
\min\!\left\{
N,\,
M_{\mathrm{threads}},\,
M_{\mathrm{memory}}
\right\},
\label{eq:local_horizon}
\\
B
&=
\left\lceil
\frac{N}{M}
\right\rceil.
\label{eq:num_blocks}
\end{align}
This adapts the horizon partition to the resources of desktop and embedded CUDA
devices while ensuring each block's working set fits in shared memory. Both
limits tighten as the augmented state grows, so \(M\) shrinks and \(B\) rises
with \(\bar{n}_x\), increasing boundary exchanges per iteration; the design
therefore favors problems whose per-stage working set is small relative to the
shared-memory budget, and a stage exceeding \(S_{\max}\) cannot be made
block-resident at all.

The resulting architecture aligns the algorithmic decomposition with the CUDA
execution and memory hierarchies: parallel-in-horizon ADMM provides stagewise
concurrency, the fused kernel removes host intervention, neighbor-local flags
bound coordination cost, and block-resident storage minimizes global-memory
traffic. Sec.~\ref{sec:tensor_framework_benchmark} measures their combined effect.

\begin{figure}[t]
    \centering
    \includegraphics[width=0.85\linewidth]{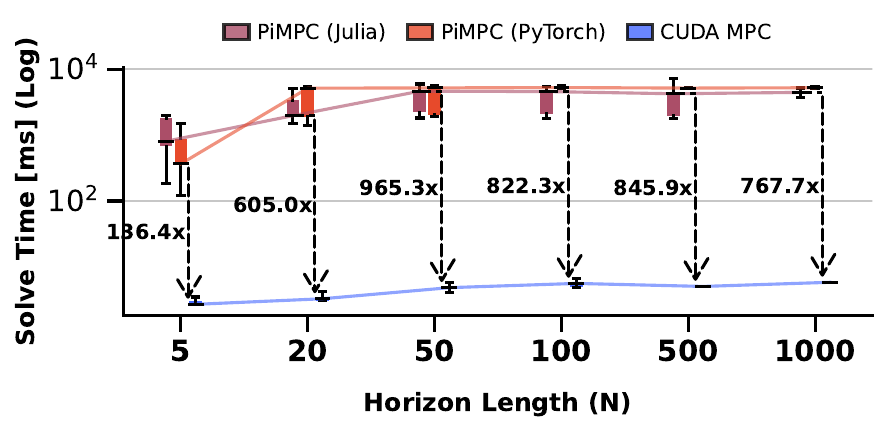}
    \includegraphics[width=0.85\linewidth]{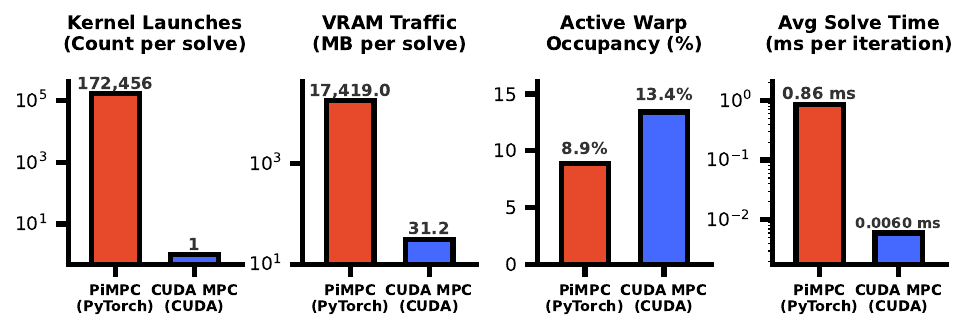}
\caption{Comparison with tensor-framework implementations of \piMPC on the
linearized AFTI-16 model.
\textbf{Top:} solve-time scaling with horizon \(N\) for \cudampc \swatch{cuda},
Julia \piMPC \swatch{pimpcjulia}, and PyTorch \piMPC \swatch{pimpcpytorch}.
\textbf{Bottom:} GPU metrics for the inner ADMM solve at \(N=1000\). Kernel
fusion reduces launches from \(172{,}456\) to \(1\), VRAM traffic from
\(17.42\)~GB to \(31.2\)~MB, and per-iteration time from \(0.86\)~ms to
\(0.0060\)~ms, while occupancy increases from \(8.9\%\) to a
shared-memory-limited \(13.4\%\).}
\label{fig:pytorch_cuda_comparison}
\end{figure}

\section{Benchmark Studies}
\label{sec:benchmarks}

We evaluate \cudampc through two benchmark studies. The first compares it with
the PyTorch and Julia implementations of \piMPC on the linearized AFTI-16
aircraft model, isolating the benefit of the GPU-native execution architecture.
The second compares it against state-of-the-art CPU and GPU MPC solvers on six
nonlinear robotics problems, measuring both solve-time scaling with horizon
length and the closed-loop performance obtained at the maximum
real-time-feasible horizon. All benchmarks use an Intel Core i7-12700H CPU,
16~GB RAM, and an NVIDIA RTX 3060 Laptop GPU with 6~GB VRAM.

\begin{figure*}[t]
    \centering
    \includegraphics[width=0.95\textwidth]{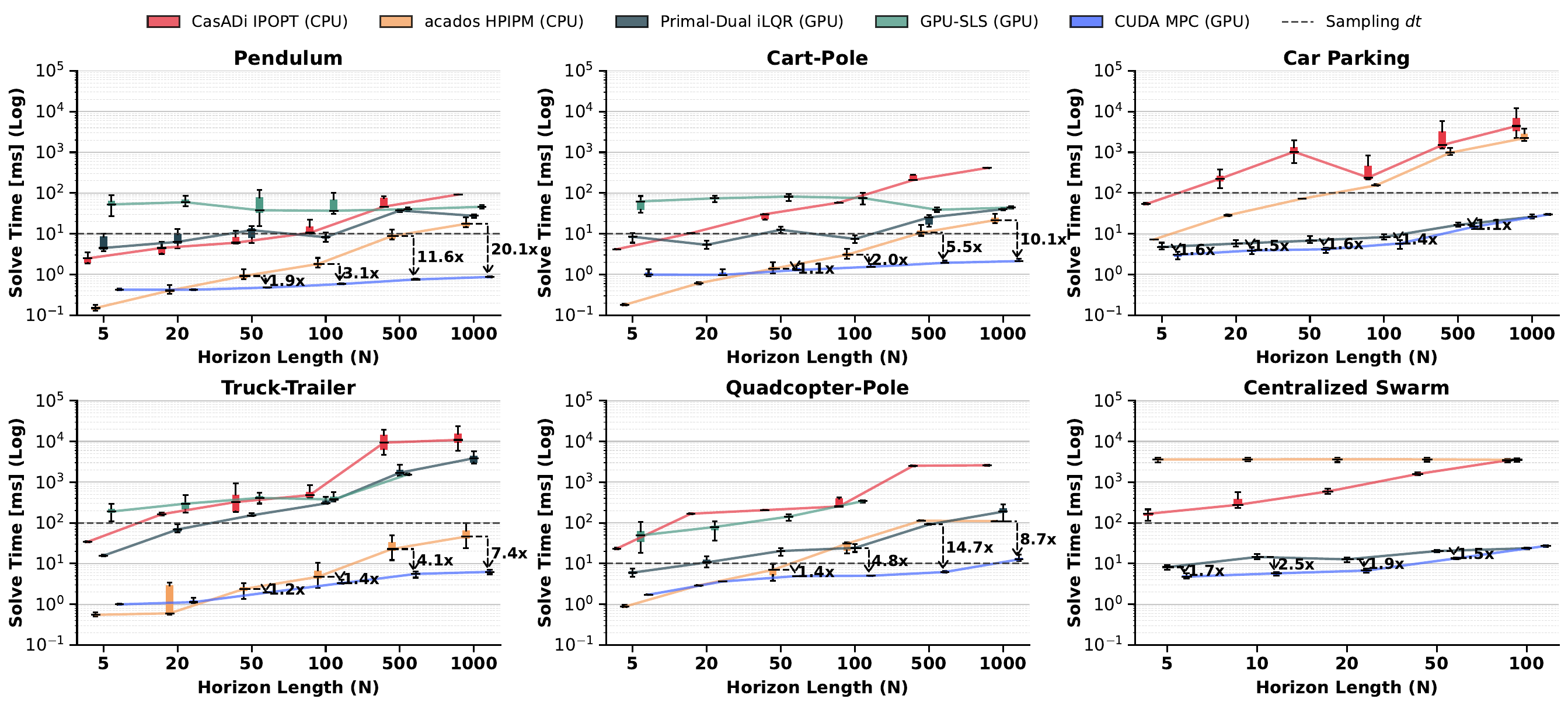}
    \caption{MPC solve-time benchmark across six nonlinear
systems---Pendulum, Cart-Pole, Car Parking, Truck-Trailer, Quadcopter-Pole, and
Centralized Swarm---for varying horizons \(N\). Box plots show logarithmic
solve-time distributions for CPU-based CasADi Ipopt \swatch{casadi} and acados
HPIPM \swatch{acados}, and GPU-based Primal-Dual iLQR \swatch{ilqr}, GPU-SLS
\swatch{sls}, and \cudampc \swatch{cuda}. Arrows indicate the speedup of \cudampc
over the next-fastest solver; the dashed line marks the real-time threshold.
GPU-SLS is omitted at larger horizons for Truck-Trailer and Quadcopter-Pole when
runs exceed the available GPU memory.}
    \label{fig:benchmark}
\end{figure*}

\subsection{Comparison with Tensor-Framework Implementations}
\label{sec:tensor_framework_benchmark}

To isolate the effect of the proposed CUDA execution architecture, we compare
\cudampc with the PyTorch and Julia implementations of \piMPC on the unstable
linearized AFTI-16 aircraft model, the numerical example used in the original
\piMPC study~\cite{wu2026pi}. All three implementations employ the same
parallel-in-horizon ADMM splitting, so the comparison isolates differences in GPU
execution, synchronization, and memory organization; our PyTorch solve times are
consistent with those reported in~\cite{wu2026pi}, confirming a comparable
baseline. We use \(dt=0.05\)~s, 2000 ADMM iterations per MPC step, and 200
simulation steps. As shown in Fig.~\ref{fig:pytorch_cuda_comparison}, \cudampc achieves a
\(136\times\) speedup at \(N=5\) and \(768\times\) at \(N=1000\), peaking at
\(965\times\) at \(N=50\). The mechanism is directly measurable per MPC solve: at \(N=1000\),
fusing the ADMM solve into one persistent kernel reduces inner-ADMM kernel
launches from \(172{,}456\) to \(1\), VRAM traffic from \(17.42\)~GB to
\(31.2\)~MB, and average iteration time from \(0.86\) to \(0.0060\)~ms. The
persistent kernel runs at low occupancy (\(13.4\%\)) because the shared-memory
workspace limits residency to one block per SM, but the solve is latency-bound
rather than throughput-bound, so this is a deliberate trade. The speedup here
reflects execution overhead alone.

\begin{table*}[t]
\caption{Maximum real-time-feasible horizon and closed-loop performance across the nonlinear benchmarks. RMSE is the root-mean-square error between the closed-loop state trajectory and the reference, in the units of each system's state. \textit{U}: unstable; \textit{L}: poor local minimum; \textit{I}: infeasible; OOT: solve time exceeds \(dt\); n.a.: formulation not expressible in the available implementation. An asterisk denotes soft barrier constraints, not directly comparable with the hard projections used by \textsc{CudaMPC}.}
\centering
\begin{tabularx}{\textwidth}{X c c c c c c c c c c c c}
\toprule
\multicolumn{1}{c}{} &
\multicolumn{3}{c}{Problem Setup} &
\multicolumn{4}{c}{CPU Solvers} &
\multicolumn{5}{c}{GPU Solvers} \\
\cmidrule(lr){2-4} \cmidrule(lr){5-8} \cmidrule(lr){9-13}

\multicolumn{1}{c}{} &
\multicolumn{3}{c}{} &
\multicolumn{2}{c}{CasADi (Ipopt) \cite{Andersson2019}} &
\multicolumn{2}{c}{acados (HPIPM) \cite{Verschueren2021}} &
\multicolumn{2}{c}{Primal-Dual iLQR~\cite{amatucci2025primal}} &
\multicolumn{1}{c}{GPU-SLS~\cite{fang2026safe}} &
\multicolumn{2}{c}{\textbf{\textsc{CudaMPC}}} \\
\cmidrule(lr){5-6} \cmidrule(lr){7-8} \cmidrule(lr){9-10} \cmidrule(lr){11-11} \cmidrule(lr){12-13}

Problem & $n_x$ & $n_u$ & $dt$ &
$N$ & RMSE & $N$ & RMSE & $N$ & RMSE & status & $N$ & RMSE \\
\midrule

Pendulum & 2 & 1 & 0.01 s & 100 & \textbf{0.860} & 500 & 0.861 & 150 & 0.862 & OOT & 1000 & 0.986 \\
Cart-Pole & 4 & 2 & 0.01 s & 20 & 30.702 (\textit{U}) & 500 & \textbf{3.364} & 100 & 21.833 (\textit{U}) & OOT & 1000 & 3.693 \\
Car Parking \cite{zhang2018autonomous} & 4 & 2 & 0.1 s & 15 & 4.749 (\textit{L}) & 60 & 4.742 (\textit{L}) & 1000 & 5.280$^{*}$ (\textit{L}) & N.A. & 1000 & \textbf{3.766} \\

Truck-Trailer \cite{altafini2001feedback} & 6 & 2 & 0.1 s & 15 & 13.661 (\textit{L}) & 1000 & \textbf{5.498} & 30 & 4.218$^{*}$ (\textit{I}) & OOT & 1000 & 6.418 \\

Quadcopter-Pole \cite{hehn2011flying} & 17 & 4 & 0.01 s & OOT & -- & 60 & -- (\textit{U}) & 5 & -- (\textit{U}) & OOT & 600 & \textbf{0.466} \\

Centralized Swarm \cite{firoozi2020distributed} & 40 & 20 & 0.1 s & OOT & -- & OOT & -- (\textit{I}) & 100 & 7.616$^{*}$ (\textit{L}) & N.A. & 100 & \textbf{4.685} \\
\bottomrule
\end{tabularx}
\label{tab:method_comparison}
\end{table*}

\subsection{Scalability and Closed-Loop Performance Across Nonlinear Benchmarks}
\label{sec:nonlinear_benchmarks}

We compare \cudampc against CasADi with Ipopt~\cite{Andersson2019} and acados
with HPIPM~\cite{Verschueren2021} on CPU, and Primal-Dual
iLQR~\cite{amatucci2025primal} and GPU-SLS~\cite{fang2026safe} on GPU, across six
nonlinear control problems: classical Pendulum and Cart-Pole setpoint regulation,
Car Parking with obstacle-avoidance constraints~\cite{zhang2018autonomous},
Truck-Trailer reversing~\cite{altafini2001feedback}, Quadcopter-Pole
stabilization~\cite{hehn2011flying}, and centralized swarm control using the 2D
formulation of~\cite{firoozi2020distributed}. Together these span increasing
state dimension, nonlinear complexity, and constraint density.

\subsubsection{Benchmark Protocol}

Solvers from different paradigms cannot be equalized by tolerance: Ipopt tests
NLP-level primal feasibility, dual feasibility and complementarity, acados
applies separate tests to its outer SQP and inner QP solves, and \cudampc's dual
variables attach to augmented-state and slack equalities rather than to the
original NLP. Mapping these onto a common KKT criterion would require
reconstructing missing multipliers and imposing shared scaling conventions, each
a method-dependent assumption.

We therefore retain each solver's default stopping criteria, standardize
iteration counts where the architectures permit, and compare computation time and
closed-loop performance in the intended real-time-control setting. All methods
use the same dynamics, objectives and constraints where supported; those
employing sequential linearization use the same number of outer iterations, and
Primal-Dual iLQR uses the default settings of~\cite{amatucci2025primal}. Reported
solve times include trajectory shifting, model linearization, matrix preparation,
the solve, and control extraction, excluding one-time JIT compilation and
graph-capture overhead. Performance is assessed through (i) solve-time scaling
with \(N\), and (ii) closed-loop performance at the largest horizon executable
within the sampling time \(dt\).

\subsubsection{Computational Scaling with Prediction Horizon}

Fig.~\ref{fig:benchmark} shows that the CPU methods remain competitive at short
horizons, particularly for low-dimensional problems, but their solve times grow
more rapidly with \(N\). The horizon-parallel structure of \cudampc, by contrast,
yields increasing benefit at longer horizons. Relative to the next-fastest
solver, its speedup grows from \(1.9\times\) at \(N=50\) to \(20.1\times\) at
\(N=1000\) for the Pendulum, and from \(1.1\times\) to \(10.1\times\) for the
Cart-Pole. At \(N=1000\), \cudampc is also \(7.4\times\) faster for the
Truck-Trailer and \(8.7\times\) faster for the Quadcopter-Pole. The arrows in
Fig.~\ref{fig:benchmark} report the margin over the \emph{next-fastest} solver,
which for several problems is another GPU method; margins over the CPU solvers
are much larger. For Car Parking at \(N=100\), \cudampc requires \(5.75\)~ms
against \(165.71\)~ms for acados---an \(28.77\times\) speedup---while the margin
over the next-fastest solver overall is \(1.4\times\).

GPU-SLS is excluded from two benchmarks for a different reason than the rest: its
available implementation cannot express the optimization-based
collision-avoidance (OBCA)~\cite{zhang2020optimization} auxiliary-variable
constraints or the pairwise inter-agent collision constraints, so Car Parking and
Centralized Swarm cannot be posed for it. Where it does run it is too slow for
the prescribed sampling intervals, reaching approximately \(0.4\)~s
(Truck-Trailer) and \(0.3\)~s (Quadcopter-Pole) at \(N=100\) before exceeding available GPU memory. Both cases are
recorded in Table~\ref{tab:method_comparison}.

\begin{figure}[t]
    \centering
    \begin{minipage}{0.47\columnwidth}
        \centering
        \includegraphics[width=\linewidth]{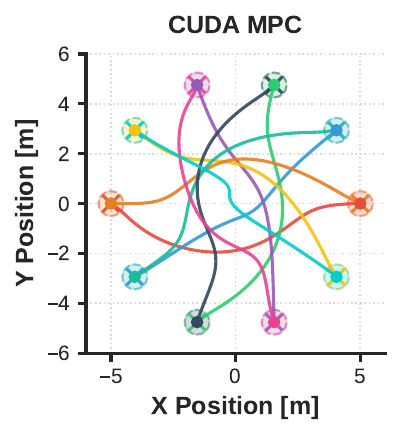}
    \end{minipage}
    \begin{minipage}{0.47\columnwidth}
        \centering
        \includegraphics[width=\linewidth]{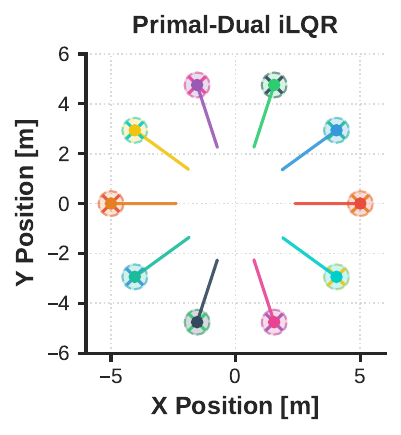}
    \end{minipage}
        \caption{Centralized swarm control with 10 agents (\(dt=0.1\)~s,
\(N=100\)). Average solve times are \(28.22\)~ms for \cudampc (left) and
\(26.37\)~ms for Primal-Dual iLQR (right). \cudampc enforces pairwise collision
constraints explicitly, whereas Primal-Dual iLQR uses soft penalties. Reducing
these penalties permits inter-agent collisions; increasing them avoids collisions
but yields the poor local minimum shown here, in which the agents deadlock.}
    \label{fig:swarm}
\end{figure}

\subsubsection{Closed-Loop Performance at the Maximum Feasible Horizon}

Table~\ref{tab:method_comparison} shows that computational speed does not
translate uniformly into improved control performance. For low-dimensional but
relatively constraint-sparse problems such as Pendulum, Cart-Pole, and
Truck-Trailer, the CPU-based second-order methods achieve performance comparable
to---and in some cases slightly better than---\cudampc despite using shorter
horizons. We hypothesize that their stronger local convergence outweighs the
benefit of additional lookahead on these problems, though the present experiments
do not isolate this effect from the accuracy attained within a fixed first-order
iteration budget.

The benefits of longer real-time-feasible horizons become more pronounced as
state dimension and constraint complexity increase. For Car Parking, \cudampc
solves the OBCA formulation with approximately \(100\,\mathrm{s}\) of lookahead
while remaining below the \(0.1\,\mathrm{s}\) sampling interval, eliminating the
separate Hybrid A* planning stage used in~\cite{zhang2018autonomous} and enabling
receding-horizon collision avoidance.

The benefits are clearest for the high-dimensional Quadcopter-Pole and swarm
problems, where longer lookahead is needed to coordinate coupled dynamics and
anticipate multiple constraints. 
For centralized control of 10 agents at $N=100$ (Fig.~\ref{fig:swarm}), \cudampc achieves an average solve time of $28.22\,\mathrm{ms}$, within the $100\,\mathrm{ms}$ sampling interval. Neither CPU solver is a viable candidate for real-time control here: acados (HPIPM) terminates at $3524.72\,\mathrm{ms}$ without finding a feasible solution. While CasADi (Ipopt) successfully returns a feasible solution, its average solve time of $4451.42\,\mathrm{ms}$ exceeds the sampling interval by more than an order of magnitude. This $\sim 4.5\,\mathrm{s}$ solve time makes CasADi (Ipopt) entirely impractical for closed-loop MPC, though its output remains useful for offline, open-loop trajectory optimization. Because neither baseline can supply a feasible solution within the required time window, we report their runtimes to highlight these practical limitations rather than as a direct speedup ratio.
Primal-Dual iLQR does run in real time, at a comparable
\(26.37\,\mathrm{ms}\), but penalties large enough to prevent collisions lead to
the deadlocked local minimum of Fig.~\ref{fig:swarm}. In this benchmark,
therefore, only \cudampc combines real-time execution with successful
collision-free coordination.

\section{Conclusion}

This paper introduced \cudampc, a GPU-native MPC solver that co-designs
parallel-in-horizon ADMM with CUDA execution and memory management. By fusing the
iterative solve into a single kernel and retaining optimization variables in
shared memory, \cudampc substantially reduces kernel-launch and global-memory
overhead, enabling long-horizon MPC at real-time control rates across nonlinear
robotic benchmarks. In the evaluated problems, longer real-time-feasible horizons
improved collision avoidance, coupled multi-agent coordination, and constraint
anticipation relative to shorter-horizon baselines. Future work will address
deployment and closed-loop validation on embedded GPU-equipped robotic platforms
under practical power, memory, and real-time computing constraints.

\bibliographystyle{IEEEtran}
\bibliography{CudaMPC_references_cleaned}

\end{document}